\documentclass[10pt,twocolumn,letterpaper]{article}
\usepackage{url}
\usepackage{iccv}
\usepackage{times}
\usepackage{epsfig}
\usepackage{graphicx}
\usepackage{amsmath}
\usepackage{amssymb}

\usepackage[breaklinks=true,bookmarks=false]{hyperref}

\iccvfinalcopy 

\def\iccvPaperID{20} 

\ificcvfinal\fi

\usepackage{amsfonts}
\DeclareMathOperator*{\argmin}{argmin}

\begin{document}

\title{End-to-end Lane Detection through Differentiable Least-Squares Fitting}
\author{Wouter Van Gansbeke \quad Bert De Brabandere \quad Davy Neven \quad Marc Proesmans \quad Luc Van Gool\\
Dept.\ ESAT, Center for Processing Speech and Images\\
KU Leuven, Belgium\\
{\tt\small \{firstname.lastname\}@esat.kuleuven.be} }

\maketitle
\ificcvfinal\thispagestyle{empty}\fi

\begin{abstract}
Lane detection is typically tackled with a two-step pipeline in which a segmentation mask of the lane markings is predicted first, and a lane line model like a parabola or spline is fitted to the post-processed mask next. The problem with such a two-step approach is that the parameters of the network are not optimized for the true task of interest (estimating the lane curvature parameters) but for a proxy task (segmenting the lane markings), resulting in sub-optimal performance. In this work, we propose a method to train a lane detector in an end-to-end manner, directly regressing the lane parameters. The architecture consists of two components: a deep network that predicts a segmentation-like \emph{weight map} for each lane line, 
and a differentiable least-squares fitting module that returns for each map the parameters of the best-fitting curve in the weighted least-squares sense. These parameters can subsequently be supervised with a loss function of choice. Our method relies on the observation that it is possible to backpropagate through a least-squares fitting procedure. This leads to an end-to-end method where the features are optimized for the true task of interest: the network implicitly learns to generate features that prevent instabilities during the model fitting step, as opposed to two-step pipelines that need to handle outliers with heuristics.
Additionally, the system is not just a black box but offers a degree of interpretability because the intermediately generated segmentation-like weight maps can be inspected and visualized. 
Code and a video is available at \url{github.com/wvangansbeke/LaneDetection_End2End}.
\end{abstract}

\section{Introduction and Related Work}
\label{sec:introduction}
A general trend in deep learning for computer vision is to incorporate prior knowledge about the problem at hand into the network architecture and loss function. Leveraging the large body of fundamental computer vision theory and recycling it into a deep learning framework gives the best of two worlds: the parameter-efficiency of engineered components combined with the power of learned features. The challenge is in reformulating these classical ideas in a manner that they can be integrated into a deep learning framework.

An illustration of this integration is at the intersection of deep learning and scene geometry~\cite{jaderberg2015spatial, yan2016perspective, handa2016gvnn, kendall2017geometric}: the general idea is to design differentiable modules that give a deep network the capability to apply geometric transformations to the data, and to use geometry-aware criterions as loss functions. The large body of classical research on geometry in computer vision~\cite{faugeras1993three, hartley2003multiple, dorst2009geometric} has inspired several methods to incorporate geometric knowledge into the network architecture~\cite{jaderberg2015spatial, yan2016perspective, handa2016gvnn, kendall2017geometric}. 
The spatial transformer network (STN) of Jaderberg~\etal~\cite{jaderberg2015spatial} and the perspective transformer net of Yan~\etal~\cite{yan2016perspective} introduce differentiable modules for the spatial manipulation of data in the network. 
Handa~\etal~\cite{handa2016gvnn} extend the STN to 3D transformations. 
Kendall~\etal~\cite{kendall2015posenet, kendall2017geometric} propose a deep learning architecture for 6-DOF camera relocalization and show that instead of naively regressing the camera pose parameters, much higher performance can be reached by taking scene geometry into account and designing a theoretically sound geometric loss function.
Other examples of exploiting geometric knowledge as a form of regularization in deep networks include~\cite{boscaini2016learning, bronstein2017geometric, kendall2017end}.

In a similar spirit, we propose in this work a lane detection method that exploits prior geometric knowledge about the task, by integrating a least-squares fitting procedure directly into the network architecture. Lane detection is traditionally tackled with a multi-stage pipeline involving separate feature extraction and model fitting steps: First, dense or sparse features are extracted from the image with a method like SIFT or SURF~\cite{lowe1999object, bay2006surf}. Second, the features are fed as input to an iterative model fitting step such as RANSAC~\cite{fischler1981random} to find the parameters of the best fitting model, which are the desired outputs of the algorithm.
In this work we replace the feature extraction step with a deep network, and we integrate the model fitting step as a differentiable module into the network. The output of the network are the lane model parameters, which we supervise with a geometric loss function.
The benefit of this end-to-end framework compared to using a multi-stage pipeline of separate steps is that all components are trained jointly: the features can adapt to the task of interest, preventing outliers during the model fitting step. Moreover, our proposed system is not just a black box but offers a degree of interpretability because the intermediately generated weight map is segmentation-like and can be inspected and visualized.

There is a vast literature on lane detection, including many recent methods that employ a CNN for the feature extraction step~\cite{wang1998lane, mccall2006video, xu2009fast, lu2015lane, gurghian2016deeplanes}. We refer to~\cite{neven2018towards} for an overview. Discussing all these approaches at length is out of the scope of this work, but most of them have in common that they tackle the task with a multi-stage pipeline involving separate feature extraction and model fitting steps. Our goal in this work is not to outperform these highly-optimized approaches, but to show that without bells and whistles, lane parameter estimation using our proposed end-to-end method outperforms a multi-step procedure.

Our work is also related to a class of methods which backpropagate through an optimization procedure. The general idea is to include an optimization process within the network itself (\textit{in-the-loop} optimization). This is possible if the optimization process is differentiable, so that the loss can be backpropagated through it~\cite{rockafellar2009variational}. One approach is to unroll a gradient descent procedure within the network~\cite{domke2012generic, metz2017unrolled}. Another approach proposed by Amos~\etal~\cite{amos2017optnet} is to solve a quadratic program (QP) problem exactly using a differentiable interior points method. Our model fitting module solves a weighted least-squares problem, which is a specific instantiation of a QP problem. Our contribution lies in showing the efficacy of including a differentiable in-network optimization step 
on a real-world computer vision task.

\section{Method}
\label{sec:method}
\begin{figure*}[t]
	\centering
	\begin{tabular}{cccc}
		\includegraphics[width=0.9\textwidth]{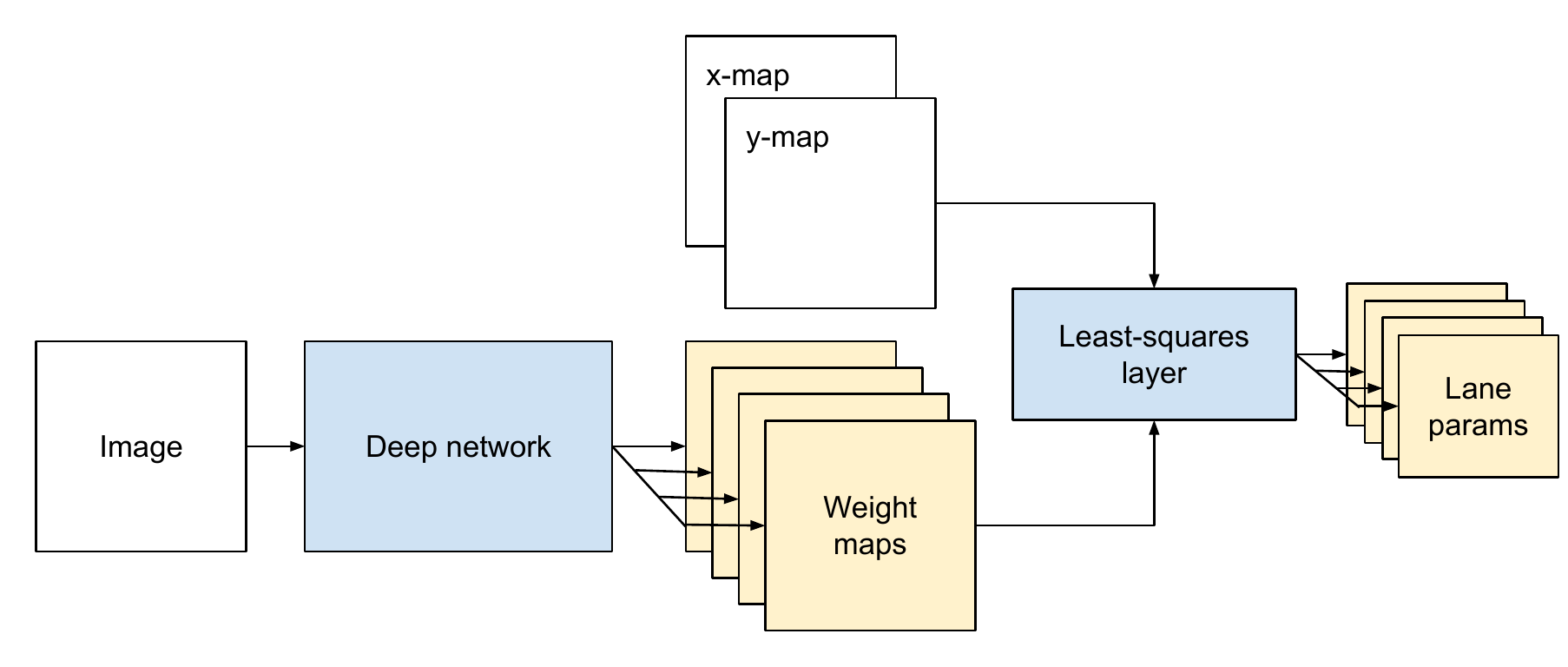}
	\end{tabular}
	\caption{\small Overview of the architecture. In this example the network produces four weight maps, each corresponding to one of four lane lines for which the parameters are estimated.}
	\label{fig:overview}
	\vspace*{-3mm}
\end{figure*}

We propose an end-to-end trainable framework for lane detection. The framework consists of three main modules, as shown schematically in Figure~\ref{fig:overview}: a deep network which generates weighted pixel coordinates, a differentiable weighted least-squares fitting module, and a geometric loss function. We now discuss each of these components in detail.

\subsection{Generating Weighted Pixel Coordinates with a Deep Network}
Each pixel in an image has a fixed $(x,y)$-coordinate associated with it in the image reference frame. In a reference frame with normalized coordinates, the coordinate of the upper left pixel is $(0,0)$ and the coordinate of the bottom right pixel is $(1,1)$. These coordinates can be represented as two fixed feature maps of the same size as the image: one containing the normalized x-coordinates of each pixel and one containing the normalized y-coordinates, indicated by the two white maps in Figure~\ref{fig:overview}.

We can equip each coordinate with a weight $w$, predicted by a deep neural network conditioned on the input image. This is achieved by designing the network to generate a feature map with the same spatial dimensions as the input image,
representing a \emph{weight} for each pixel coordinate. Any off-the-shelf dense prediction architecture can be used for this. In order to restrict the weight maps to be non-negative, the output of the network is squared. If we flatten the generated weight map and the two coordinate maps, we obtain a list of $m$ triplets $(x_i,y_i,w_i)$ of respectively the x-coordinate, y-coordinate and coordinate weight for each pixel $i$ in the image. For an image of height $h$ and width $w$, the list contains $m = h \cdot w$ triplets. This list is the input to the weighted least-squares fitting module discussed next.

For the task of lane detection, the network must generate multiple weight maps: one for each lane line that needs to be detected. 
A lane line (also referred to as \textit{curve}) is defined as the line that separates two lanes, usually indicated by lane markings on the road. E.g. in the case of ego-lane detection, the network outputs two weight maps; one for the lane line immediately to the left of the car and one for the lane line immediately to the right, as these lines constitute the borders of the ego-lane.

\subsection{Weighted Least-Squares Fitting Module}
The fitting module takes the list of $m$ triplets $(x_i,y_i,w_i)$ and interprets them as weighted points in 2D space. Its purpose is to fit a curve (e.g. a parabola, spline or other polynomial curve) 
through the list of coordinates in the weighted least-squares sense, and to output the $n$ parameters of that best-fitting curve. 

\paragraph{Least-squares fitting.} Many traditional computer vision methods employ curve fitting as a crucial step in their pipeline. One fundamental and simple fitting procedure is linear least-squares. Consider a system of linear equations 
\[ X \beta = Y \] 
with $X \in \mathbb{R}^{m \times n}$, $\beta \in \mathbb{R}^{n \times 1}$, and $Y \in \mathbb{R}^{m \times 1}$:
\[
X = 
\begin{pmatrix}
1 & x_1 & \cdots & x_1^{n-1} \\
1 & x_2 & \cdots & x_2^{n-1} \\
\vdots  & \vdots  & \ddots & \vdots  \\
1 & x_m & \cdots & x_m^{n-1} 
\end{pmatrix}, \: \beta = \begin{pmatrix}
\beta_{1}  \\
\beta_{2} \\
\vdots \\
\beta_{n}
\end{pmatrix}, \: Y = \begin{pmatrix}
y_{1}  \\
y_{2} \\
\vdots \\
y_{m}
\end{pmatrix}.
\]

There are $m$ equations in $n$ unknowns. If $m > n$, the system is overdetermined and no exact solution exists. We resort to finding the least-squares solution, which
minimizes the sum of squared differences between the data values and their corresponding modeled values:
\begin{equation}
\beta = \argmin_{Z \in \mathbb{R}^{n \times 1}} ||X Z - Y||^2.
\end{equation}
The solution is found by solving the normal equations, and involves the matrix multiplication of $Y$ with the pseudo-inverse of $X$:
\begin{equation}
\label{eq:pseudoinverse}
\beta = \left( X^T X \right)^{-1} X^T Y.
\end{equation}

\paragraph{Weighted least-squares fitting.}
The previous formulation can be extended to a \emph{weighted} least-squares problem. Let $W \in \mathbb{R}^{m \times m}$ be a diagonal matrix containing weights $w_i$ for each observation. In our framework, the observation will correspond to the fixed $(x,y)$-coordinates in the image reference frame, and the weights will be generated by a deep network conditioned on the image. The weighted least-squares problem is 
\begin{equation}
\label{eq:weightedlq}
W X \beta = W Y.
\end{equation}
By defining $X' = WX$ and $Y' = WY$ with 
\[
W =
\mathrm{diag(}
\left[
\begin{array}{c} 
w_1 \\ w_2 \\ \vdots \\ w_{m}
\end{array} 
\right]
\mathrm{)}
=
\begin{pmatrix}
w_{1} & 0 & \cdots & 0 \\
0 & w_{2} & \cdots & 0 \\
\vdots  & \vdots  & \ddots & \vdots  \\
0 & 0 & \cdots & w_{m}
\end{pmatrix},
\]
we can reformulate this in the standard form $X' \beta = Y'$ and solve it in the same way as before.

\paragraph{Backpropagating through the fitting procedure.}
Recall that we have a list of weighted pixel coordinates $(x_i,y_i,w_i)$ where the coordinates $(x_i,y_i)$ are fixed and the weights $w_i$ are generated by a deep network conditioned on an input image. We can use these values to construct the matrices $X$, $Y$ and $W$, solve the weighted least-squares problem, and obtain the parameters $\beta$ of the best-fitting curve through the weighted pixel coordinates.

The contribution of this work lies in the following insight: instead of treating the fitting procedure as a separate post-processing step, we can \emph{backpropagate through it} and apply a loss function on the parameters of interest $\beta$ rather than indirectly on the weight maps produced by the network. This way, we obtain a powerful tool for tackling lane detection within a deep learning framework in an end-to-end manner.

Note that equations~\ref{eq:pseudoinverse} and ~\ref{eq:weightedlq} only involve differentiable matrix operations. It is thus possible to calculate derivatives of $\beta$ with respect to $W$, and consequently also with respect to the parameters of the deep network. The specifics of backpropagating through matrix transformations are well understood. We refer to~\cite{giles2008extended} for the derivation of the gradients of this problem using Cholesky decomposition. 

By backpropagating the loss through the weighted least-squares problem, the deep network can learn to generate a weight map that gives \emph{accurate lane line parameters} when fitted with a least-squares procedure, rather than optimizing the weight map for a proxy objective like \emph{lane line segmentation} with curve fitting as a separate post-processing step.

The model parameters that are the output of the curve fitting step can be supervised directly with a mean squared error criterion or via a more principled geometric loss function as the one discussed in the next section.

\subsection{Geometric Loss Function}
\label{sec:geometric_loss}

\begin{figure}[t]
	\centering
	\includegraphics[width=0.4\textwidth]{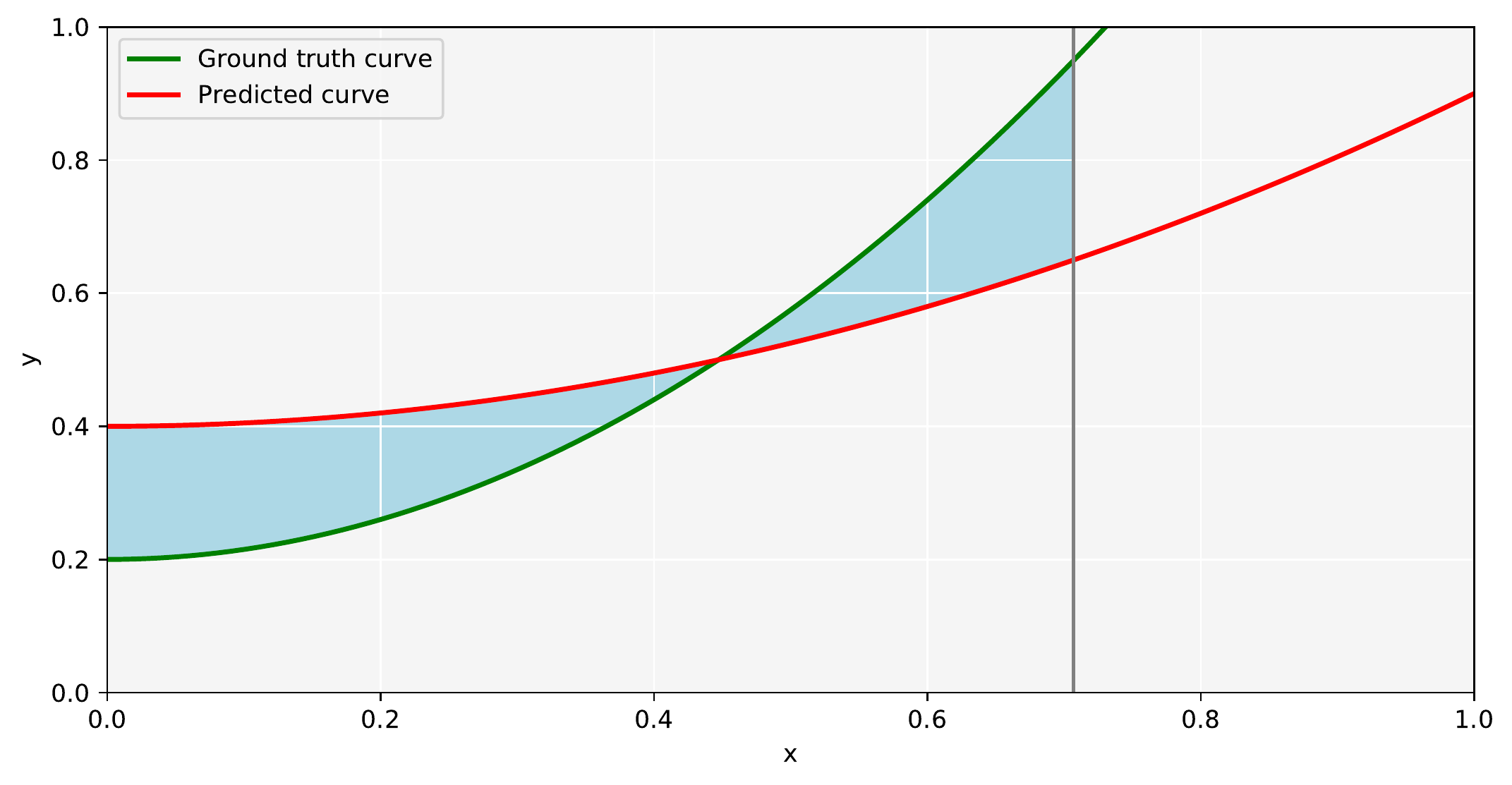}
	\caption{The geometric loss minimizes the (squared) area between the predicted curve and ground truth curve up to a point $t$.}
	\label{fig:area_loss}
\end{figure}

The $n$ curve parameters $\beta_i$ that are the output of the curve fitting step could be supervised by comparing them with the ground truth parameters $\hat{\beta}_i$ with a mean squared error criterion, leading to the following L2 loss:
\begin{equation}
L = \frac{1}{n}\sum\limits_{i=1}^n (\beta_i - \hat{\beta}_i)^2.
\end{equation}
The problem with this is that the curve parameters have different sensitivities: a small error in one parameter value might have a larger effect on the curve shape than an error of the same magnitude in another parameter.

Ultimately, the design of a loss function depends on the task: it must optimize a relevant metric for the task of interest. For lane detection, we opt for a loss function that has a geometric interpretation: it minimizes the squared area between the predicted curve $y_{\beta}(x)$ and the ground truth curve $y_{\hat{\beta}}(x)$ in the image plane, up to a point $t$ (see Figure~\ref{fig:area_loss}):
\begin{equation}
L = \int_{0}^{t} (y_{\beta}(x) - y_{\hat{\beta}}(x))^2 dx
\end{equation}

For a straight line $y = \beta_0 + \beta_1 x$, this results in:
\begin{equation}
L = \Delta \beta_0^2t + \Delta \beta_1\Delta \beta_0t^2 + \frac{\Delta \beta_1^2t^3}{3},
\end{equation}
where $\Delta \beta_i = \beta_i - \hat{\beta}_i $. For a parabolic curve $y = \beta_0 + \beta_1 x + \beta_2 x^2$ it gives:
\begin{equation}
\begin{aligned}
L = \frac{\Delta \beta_2^2t^5}{5}+\frac{2\Delta \beta_2\Delta \beta_1t^4}{4}+\frac{(\Delta \beta_1^2+2 \Delta \beta_2\Delta \beta_0)t^3}{3} \\  +\frac{2\Delta \beta_1\Delta \beta_0t^2}{2}+\Delta \beta_0^2t.
\end{aligned}
\end{equation}

\subsection{Optional Transformation to Other Reference Frame}
Before feeding the list of weighted coordinates $(x_i,y_i,w_i)$ to the fitting module, the coordinates can optionally be transformed to another reference frame by multiplying them with a transformation matrix $H$. This multiplication is also a differentiable operation through which backpropagation is possible.

In lane detection, for example, a lane line is better approximated as a parabola in the ortho-view (i.e., the top-down view) than as a parabola in the original image reference frame. We can achieve this by simply transforming the coordinates from the image reference frame to the ortho-view by multiplying them with a homography matrix $H$. The homography matrix is considered known in this case. Note that it is not the input image that is transformed to the ortho-view, only the list of coordinates.

\section{Experiments}
\label{sec:experiments}
To better understand the dynamics of backpropagating through a least-squares fitting procedure, we first provide a simple toy experiment. Next, we evaluate out method on the real-world task of lane detection. Our goal is not to extensively tune the method and equip it with bells and whistles to reach state-of-the-art performance on a lane detection benchmark, but to illustrate that end-to-end training with our method outperforms the classical two-step procedure in a fair comparison.
\subsection{Toy Experiment}
\label{sec:toy_experiment}
\begin{figure*}[t]
	\centering
	\includegraphics[width=0.9\textwidth]{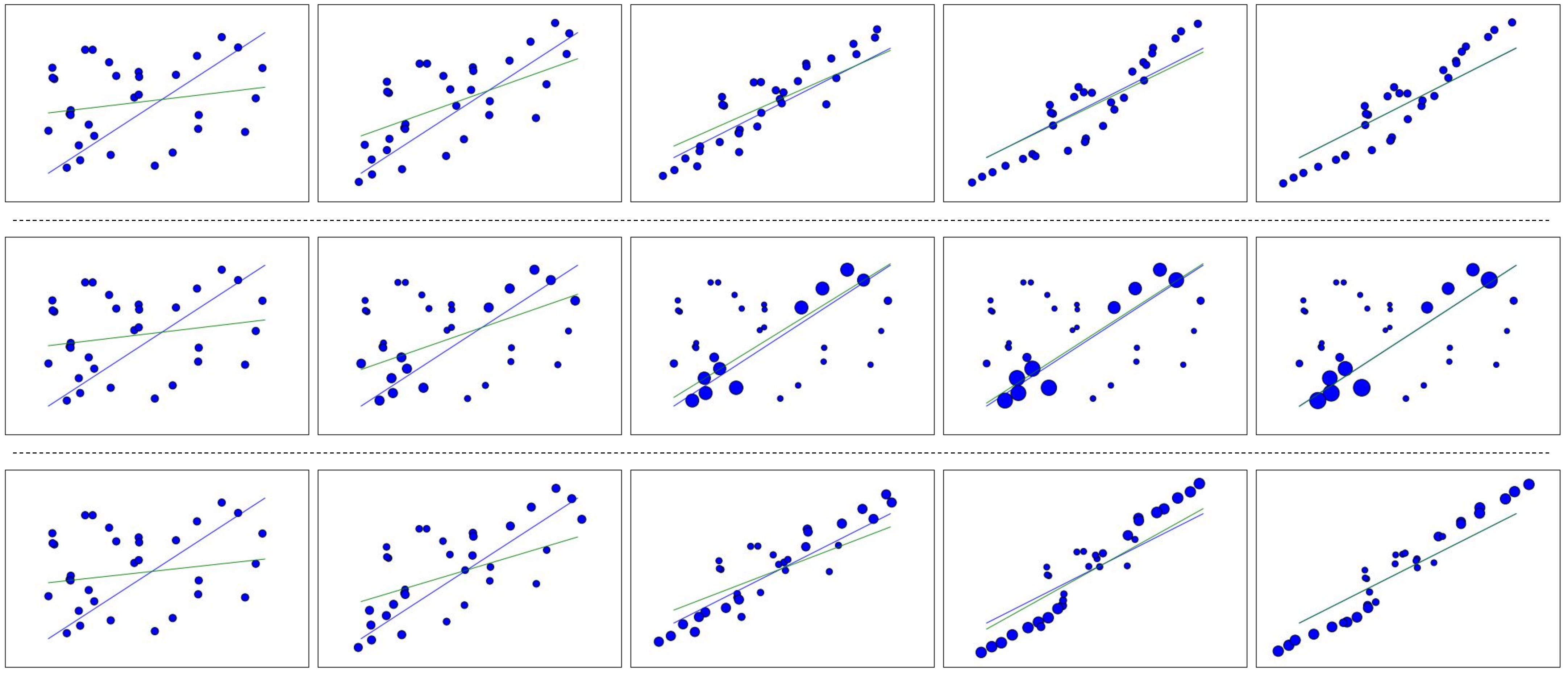}
	\caption{Illustration of differentiable weighted least-squares line fitting on a synthetic example, explained in Section~\ref{sec:toy_experiment}.}
	\label{fig:weighted_least_squares}
\end{figure*}

Recall that the least-squares fitting module takes a list of weighted coordinates $(x_i,y_i,w_i)$ as input and produces the parameters $\beta_i$ of the best-fitting curve (e.g. a polynomial) through these points as output. During training, the predicted curve is compared to the ground truth curve (e.g. an annotated lane) in the loss function, and the loss is backpropagated to the inputs of the module. Note that we only discussed backpropagating the loss to the \emph{coordinate weights}, generated by a deep network, but that it is in principle also possible to backpropagate the loss to the coordinates themselves.

This is illustrated in Figure~\ref{fig:weighted_least_squares}. The blue dots represent coordinates and their size represents their weight. The blue line is the best-fitting straight line (i.e., first-order polynomial) through the blue dots, in the weighted least-squares sense. The green line represents the ground truth line, and is the target. If we design the loss as in Section~\ref{sec:geometric_loss} and backpropagate through the fitting module, we can iteratively minimize the loss through gradient descent such that the predicted line converges towards the target line. This can happen in three ways: 
\begin{enumerate}
	\item By updating the coordinates $(x_i, y_i)$ while keeping their weights $w_i$ fixed. This corresponds to the first row, where the blue dots move around but keep the same size.
	\item By updating the weights $w_i$ while keeping the coordinates $(x_i, y_i)$ fixed. This corresponds to the second row, where the blue dots change size but stay at the same spot.
	\item By updating both the coordinates $(x_i, y_i)$ and the weights $w_i$. This corresponds to the third row, where the blue dots move around and change size at the same time.
\end{enumerate}

For the lane detection task in the next section we focus on the second case, where the coordinate locations are fixed, as they represent image pixel coordinates that lie on a regular grid. The loss is thus only backpropagated to the coordinate \emph{weights}, and from there further into the network that generates them, conditioned on an input image.

\subsection{Ego-lane Detection}
We now turn to the real-world task of ego-lane detection. To be more precise, the task is to predict the parameters of the two border lines of the ego-lane (i.e. the lane the car is driving in) in an image recorded from the front-facing camera in a car. As discussed before, the traditional way of tackling this task is with a two-step pipeline in which features are detected first, and a lane line model is fitted to these features second. The lane line model is typically a polynomial or spline curve. In this experiment, the lines are modeled as parabolic curves $y=ax^2+bx+c$ in the ortho-view. The network must predict the parameters $a, b$ and $c$ of each curve from the untransformed input image. The error is measured as the normalized area between the predicted curve $y(x)$ and the ground truth curve $\hat{y}(x)$ in the ortho-view up to a fixed distance $t$ from the car:
\begin{equation}
\textrm{error} = \int_{0}^{t} |y(x)-\hat{y}(x)| dx 
\end{equation}
It is averaged over the two lane lines of the ego-lane and over the images in the dataset.

Again, it is not our goal to outperform the sophisticated and highly-tuned lane detection frameworks found in the literature, but rather to provide an apples-to-apples comparison of our proposed end-to-end method to a classical two-step pipeline. Extensions like data augmentation, more realistic lane line models, and an optimized base network architecture are orthogonal to our approach. In order to provide a fair comparison, we train the same network in two different ways, and measure its performance according to the error metric. We compare following two methods:

\begin{description}

\item[Cross-entropy training] In this setting, the network that generates the pixel coordinate weights (two weight maps: one for each lane line) is trained in a segmentation-like manner, with the standard per-pixel binary cross-entropy loss. This corresponds to the feature detection step in a two-step pipeline. The segmentation labels are created from the ground truth curve parameters by drawing the corresponding curve with a fixed thickness as a dense label. At test time, a parabola is fitted through the predicted features in the least-squares sense. This corresponds to the fitting step in a two-step pipeline.
\item[End-to-end training (ours)] In this setting, the network is trained with our proposed method involving backpropagation through a weighted least-squares fit and the geometric loss function. There is no need to create any proxy segmentation labels, as the supervision is directly on the curve parameters. It corresponds to the second case in Section~\ref{sec:toy_experiment}.
\end{description}
\begin{figure*}[t]
    \begin{center}
	\begin{tabular}{cc}
		{\includegraphics[width=7.0cm]{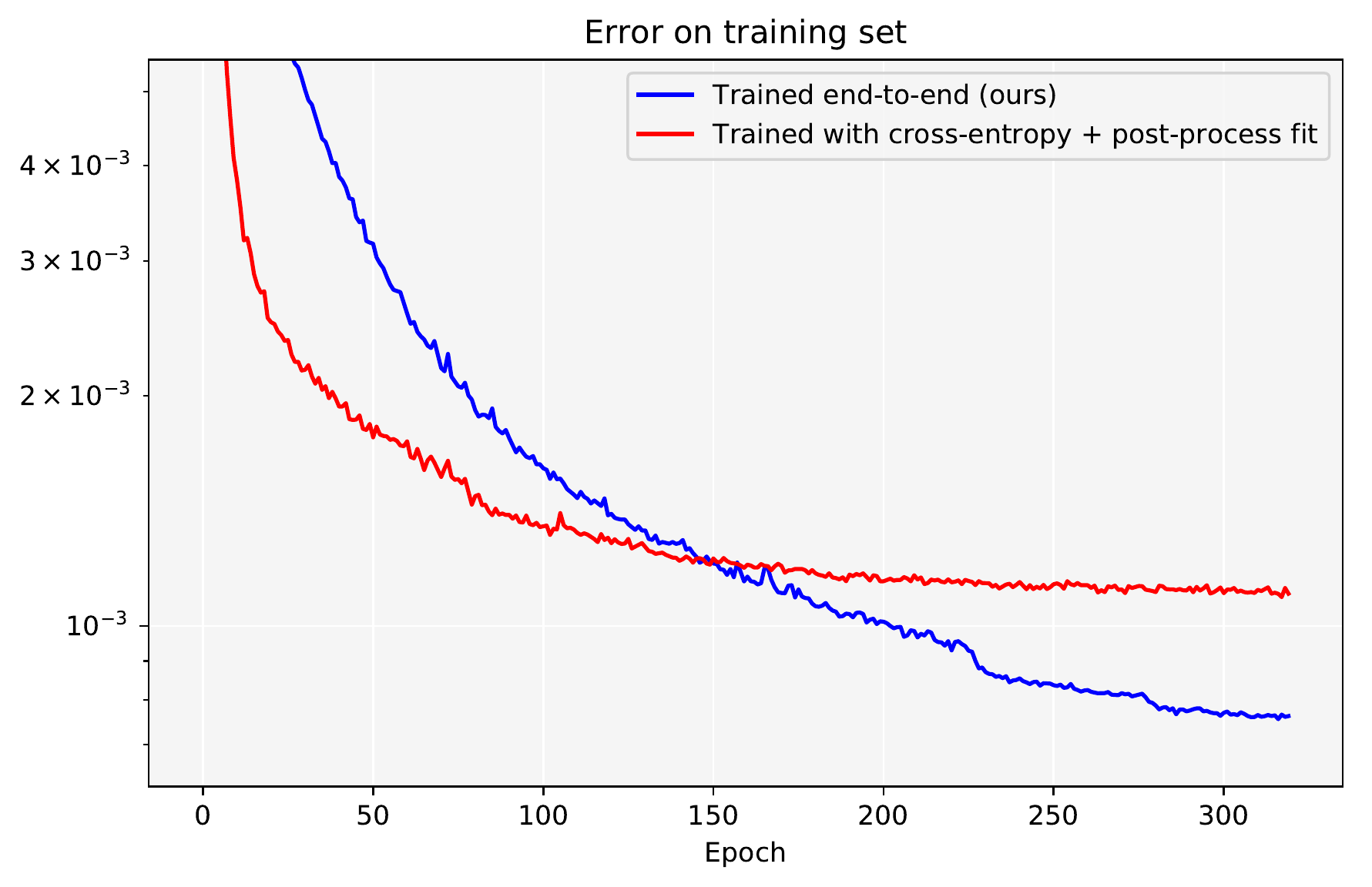}}&
	    {\includegraphics[width=7.0cm]{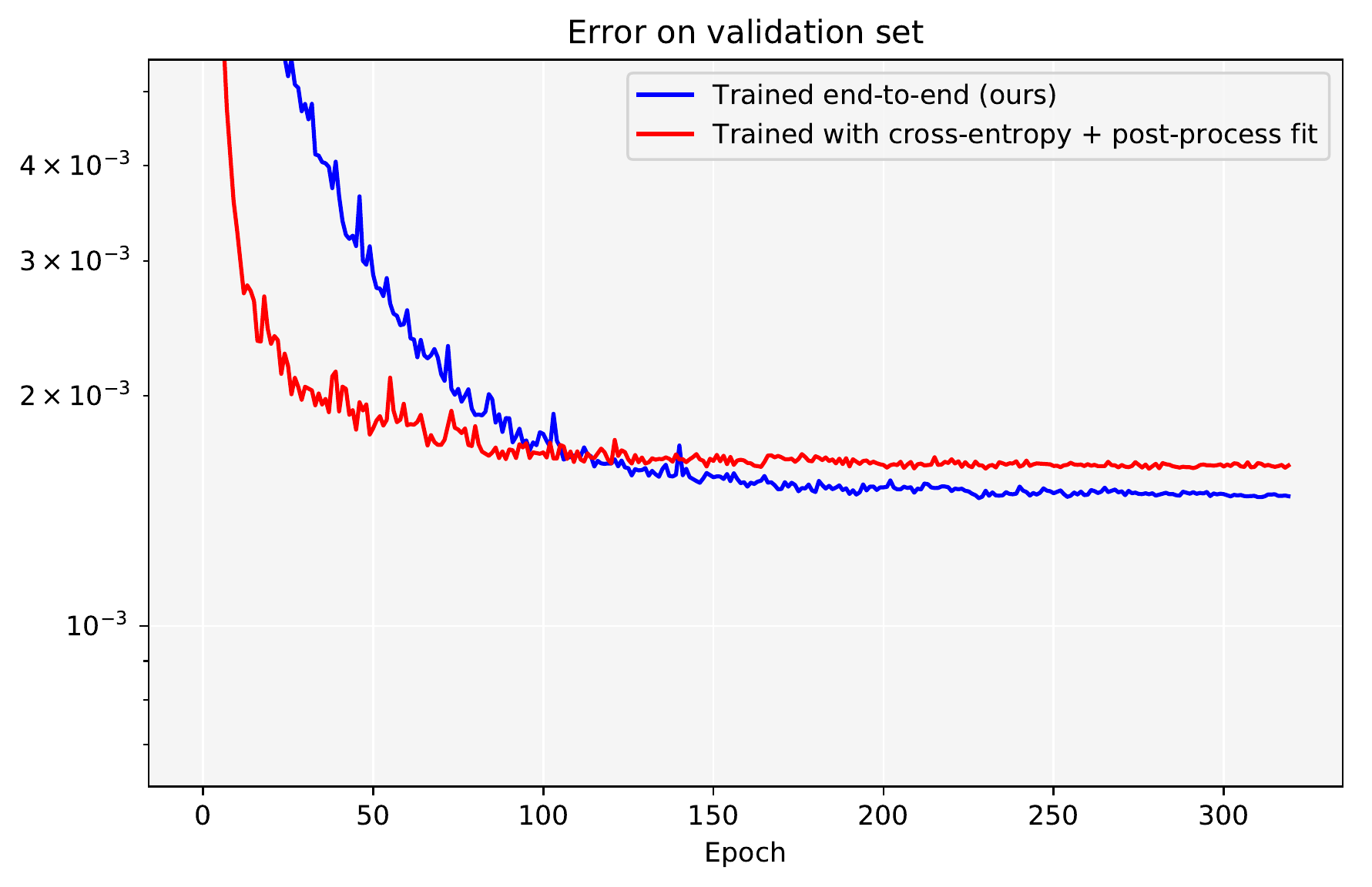}}\\
	\end{tabular}
	\caption{Convergence of the error during training (left) on the training set and (right) on the validation set. The error curve of our method is indicated in blue, the error curve of the method trained with a segmentation-like cross-entropy loss and fitting as a post-processing step is indicated in red.}
	\label{fig:results}
	\end{center}
\end{figure*}

\begin{table}
	\caption{Summarized results of the experiment. Our end-to-end method gives lower error on both the training and validation set.}
	\begin{center}
		\begin{tabular}{lll||r|r}
			&\multicolumn{2}{p{1.5cm}||}{~}&
			\multicolumn{2}{c}{\centering Metric}\\
			& \multicolumn{2}{p{1.5cm}||}{\centering Method} &
			\multicolumn{1}{c|}{\centering loss} &
			\multicolumn{1}{c}{\centering error}\\
			\hline
			Train & \multicolumn{2}{l||}{Cross-entropy} & - & 1.113e-3 \\
			& \multicolumn{2}{l||}{End-to-end (ours)} & 7.340e-7 & \textbf{7.590e-4} \\
			\hline
			Val & \multicolumn{2}{l||}{Cross-entropy} & - & 1.603e-3 \\
			& \multicolumn{2}{l||}{End-to-end (ours)} & 1.912e-5 & \textbf{1.437e-3} \\
		\end{tabular}
	\end{center}
	\label{tab:results}
	\vspace*{-5mm}
\end{table}

\paragraph{Dataset and training setup}
We run our experiment on the TuSimple dataset~\cite{tusimple}. We manually select and clean up the annotations of 2535 images, filtering out images where the ego-lane cannot be detected unambiguously.
20\% of the images are held out for validation, taking care not to include images of a single temporal sequence in both train and val set.

ERFNet~\cite{romera2017efficient} is used as the network architecture. The last layer is adapted to output two feature maps, one for each ego-lane line. In both the cross-entropy and end-to-end experiments, we train for 350 epochs on a single GPU with image resolution of 256x512, batch size of 8, and Adam~\cite{kingma2015adam} with a learning rate of 1e-4. As a simple data augmentation technique the images are randomly flipped horizontally. In the end-to-end experiments, we use a fixed transformation matrix H to transform the weighted pixel coordinates to the ortho-view. 
Note that the input image itself is not transformed to the ortho-view, although that would also be an option. 
The system is implemented in PyTorch~\cite{paszke2017automatic}.

\paragraph{Results and discussion}
Figure~\ref{fig:results} shows the error (i.e., normalized area between gt and predicted curve) during training for both methods, on the training set (left) and on the validation set (right). Table~\ref{tab:results} summarizes these results and reports the value of the geometric loss (see Section~\ref{sec:geometric_loss}), which is the actual metric being optimized in our end-to-end method.

\begin{figure*}[t]
	\centering
	\includegraphics[width=0.9\textwidth]{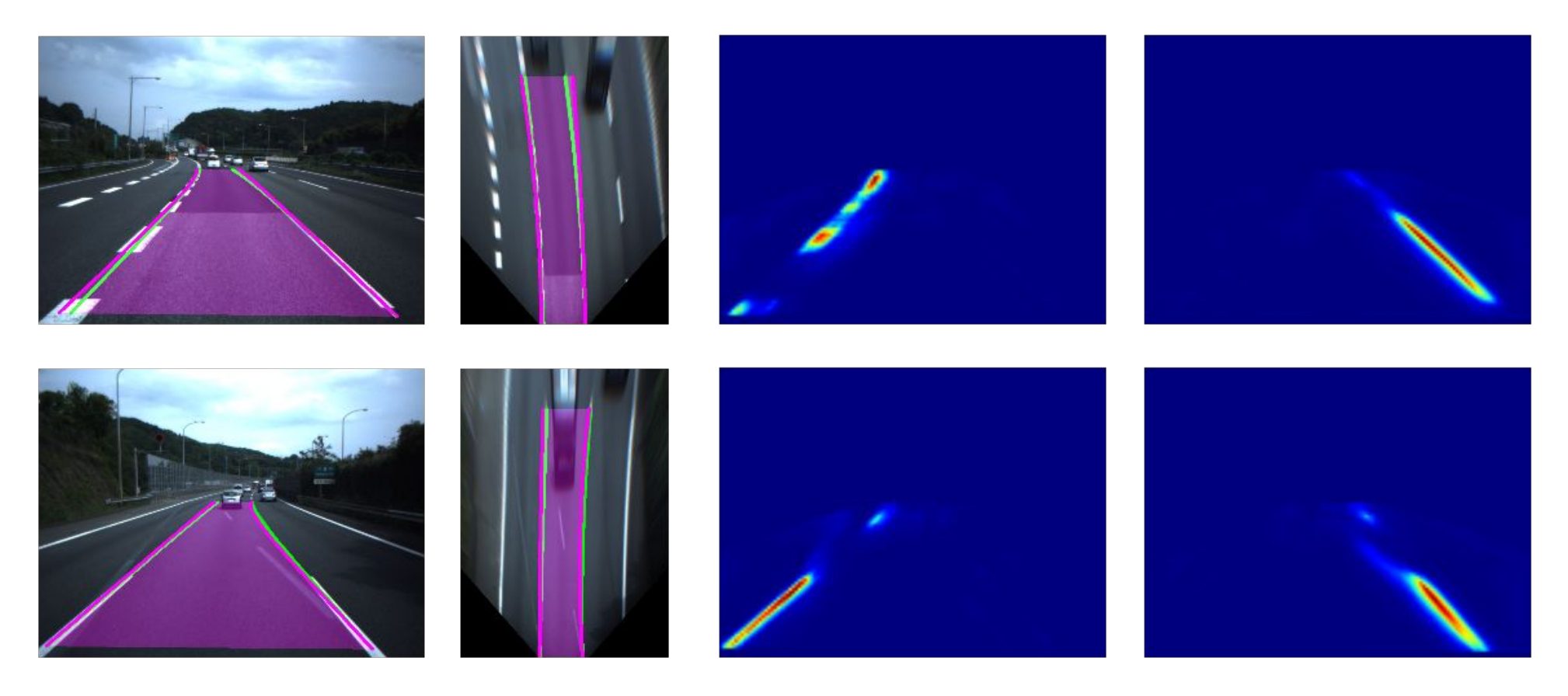}
	\caption{Qualitative results on the lane detection task. From left to right: input image with overlayed ground truth (green) and predicted lane lines (purple), ortho-view of the scene in which the loss is calculated (see text), coordinate weight maps corresponding to left and right lane lines. The network learns to handle the large variance in lane markings and challenging conditions like the faded markings in the bottom row that are correctly ignored.
	}
	\label{fig:lane_detection}
\end{figure*}

We see that our end-to-end method converges to lower error than the method trained with cross-entropy loss, both on the training and validation set. 
The convergence is slower, but this should come as no surprise: the supervision signal in the end-to-end method is much weaker than in the cross-entropy method with dense per-pixel labels. To see this, consider that the end-to-end method does not explicitly force the weight map to be a segmentation of the actual lane lines in the image. Even if the network generated a seemingly random-looking weight map, the loss (and thus the gradients) would still be zero as long as the least-squares fit through the weighted coordinates would coincidentally correspond to the ground truth curve. For example, the network could fall into a local minimum of generating the weight map based on image features such as the vanishing point at the horizon and the left corner of the image, still resulting in a relatively well fitting curve but hard to improve upon. One option to combine the fast convergence of the cross-entropy method with the superior performance of the end-to-end method would be to pre-train the network with the first and fine-tune with the latter.

Figure~\ref{fig:lane_detection} shows some qualitative results of our method. Despite the weak supervision signal, the network eventually discovers that the most consistent way to satisfy the loss function is to focus on the visible lane markings in the image, and to map them to a segmentation-like representation in the weight maps. The network learns to handle the large variance in lane markings and can tackle challenging conditions like the faded markings in the bottom example of Figure~\ref{fig:lane_detection}. 

To be robust against outliers in the fitting step, classic methods often resort to iterative optimization procedures like RANSAC~\cite{fischler1981random}, iteratively reweighted least-squares~\cite{holland1977robust} and iterative closest point~\cite{besl1992method}. In our end-to-end framework, the network \emph{learns} a mapping from input image to weight map such that the fitting step becomes robust. This moves complexity from the post-processing step into the network, allowing for a simple one-shot fitting step.

\subsection{Multi-lane Detection}
As a final experiment, we extend our method to handle multi-lane detection and compare with state-of-art. Now, the goal is to detect the ego-lanes aswell as the farther lane-lines in the driving scene (i.e. 4 lane lines in total). The framework must predict 4 sets of coefficients $a, b$ and $c$ to model the four lane lines accurately in the ortho-view. This means that the network has four output maps. 

\begin{figure*}[t]
	\centering
	\includegraphics[width=0.9\textwidth]{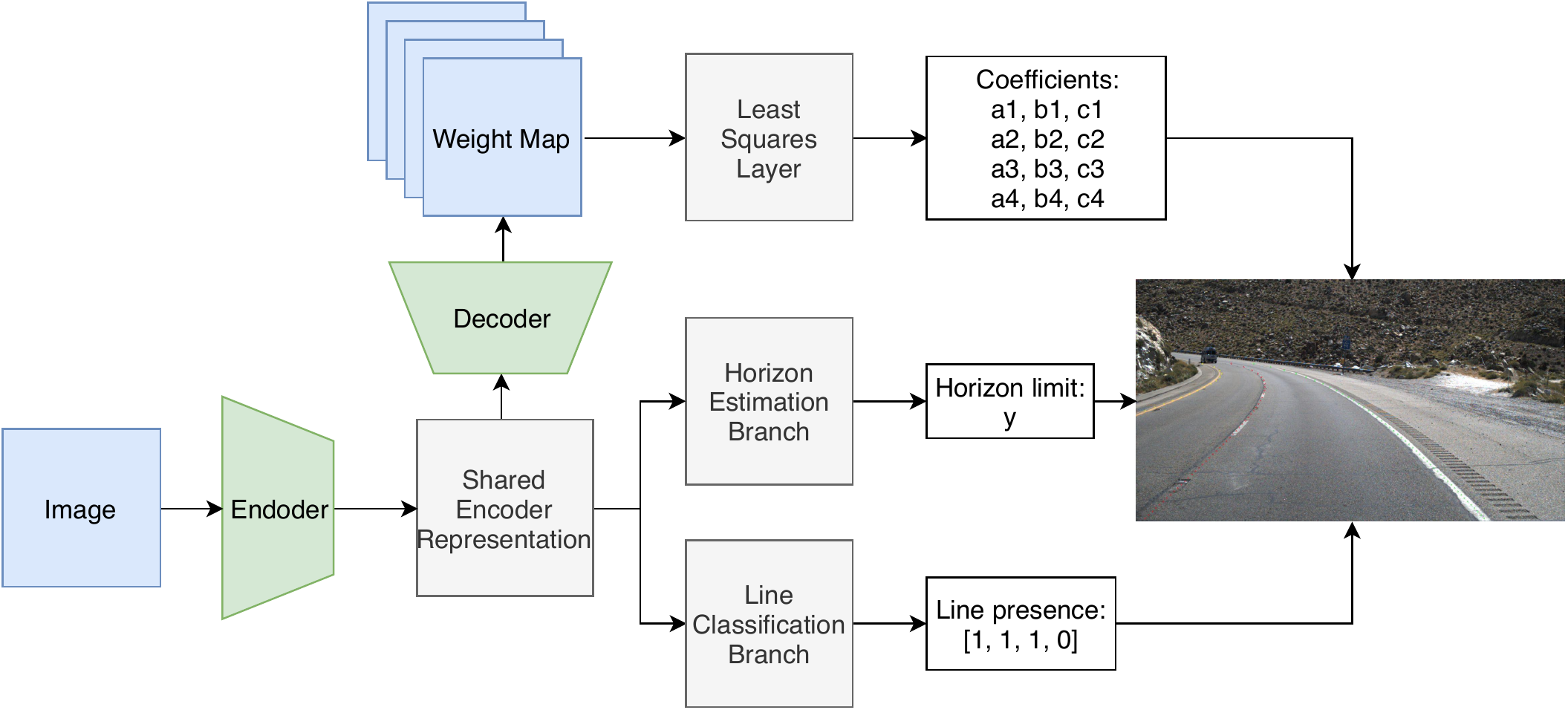}
	\caption{The framework for our multi-lane detection experiment is shown here. The three tasks are represented as the three branches in the figure which are necessary to generate the final line coordinates. Notice that the line classification branch informs the framework that there are only three lane lines in the image. This controls the amount of false positives.
	}
	\label{fig:multi_lane_fig}
\end{figure*}

In order to get the final coordinates of the lane lines, we augment our method in a shared encoder architecture with a line prediction branch and a horizon prediction branch. Since we output all the coefficients for a fixed amount of lane lines, we ought to know if the predicted line is present and where the line ends. The TuSimple benchmark requires (x, y)-coordinates for the evaluation after all. We argue that these additional tasks can benefit from the representations learned by the main task. By adopting a multi-task framework we don't unnecessarily decrease the speed of the framework since we can share the encoder. The added branches are kept straightforward by using 4 convolution layers with 3x3 kernel size, followed by max pooling and finally a fully connected layer. The horizon estimation branch performs a regression to estimate the horizon and the line classification branch determines if a line is present or not. This framework is visualized in Figure~\ref{fig:multi_lane_fig}. The losses for all tasks are linearly combined before backpropagation and we train the whole framework completely end-to-end. Furthermore, no segmentation masks are used during training. More details about the architecture can be found in the code. 

\paragraph{Dataset and training setup}
We use the TuSimple dataset to validate the effectiveness of our method for multi-lane detection. Here, we use the complete 3626 images from which we also leave 20\% out for validation. The testset consists of 2782 images. We adopt the same settings as in the ego-lane detection experiment. However, instead of using the pseudoinverse of $X$, a Cholesky decomposition is used in order to be less sensitive to ill-conditioned matrices when solving a system of linear equations. 

\begin{figure}[t]
	\centering
	\includegraphics[width=0.3\textwidth]{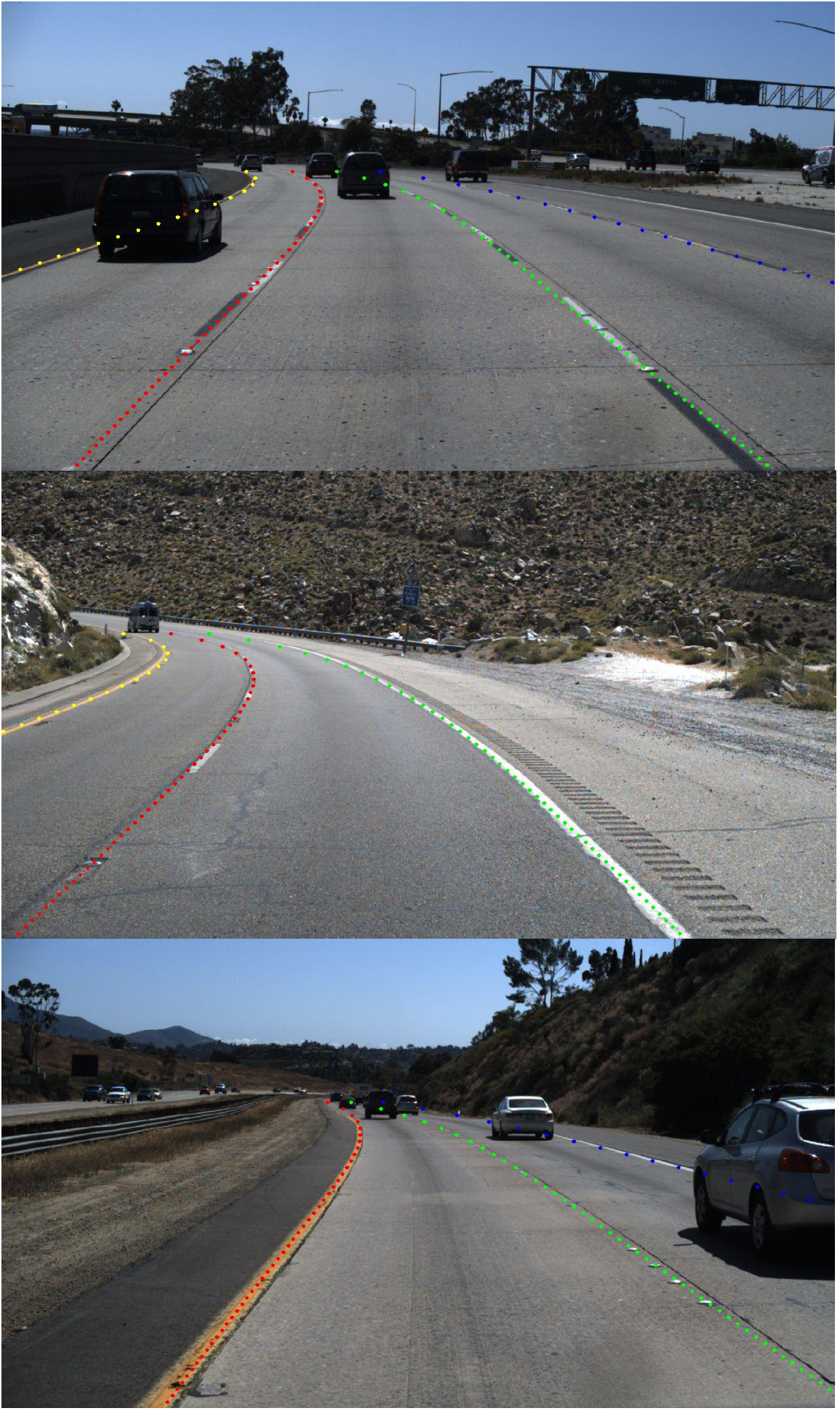}
	\caption{Predicted curves on the testset of TuSimple.
	}
	\label{fig:pred}
\end{figure}

\begin{table}
	\caption{Comparison with state-of-the-art.}
	\begin{center}
		\begin{tabular}{l||c|c|c}
		     {\centering Method} & {\centering Acc} & {\centering Fps}& {\centering Extra data}\\
			\hline
		    XingangPan~\cite{pan2018}~\ & 96.53\% & 5.51 & yes\\
			DavyNeven~\cite{neven2018towards} & 96.40\% & 52.6 & no \\
			\textbf{Ours} & 95.80\% & 71.5 & no\\
			\hline
			Baseline  & 95.10\% & N.A. & no\\
		\end{tabular}
	\end{center}
	\label{tab:sota}
	\vspace*{-5mm}
\end{table}
\paragraph{Results and Discussion}
We first compare to our baseline: the conventional two step approach, i.e. segmentation followed by a line fitting module. We used ERFNet trained in the same way as before but without backpropagation through our least squares layer. 
The results in table~\ref{tab:sota} show that we clearly outperform this baseline. The accuracy is calculated as the ratio of the number of correct points to the number of points in the ground truth. We improve by 0.7\% when training end-to-end. Figure~\ref{fig:pred} shows some predictions on the testset. 
The baseline is more sensitive towards outliers since the line fitting is independent from the optimization process. Our end-to-end approach allows us to penalize those occurrences while optimizing for the final coordinates. We also show that our fairly elegant method is not far off from the leading methods on the TuSimple benchmark. Furthermore, our method is considerably faster (71 fps on a NVIDIA 1080Ti) since we don't require any post-processing to predict the line coordinates. It also does not require expensive segmentation ground truth. This proves that the network can learn the coefficients for multi-lane detection jointly by backpropagation through our differentiable weighted least squares layer.
A restriction is that the network outputs a fixed amount of weight maps with a predefined order. In this way, lane changes are hard to handle since the order is ambiguous during this event. Each output map corresponds to a weighted least squares problem for a specific lane line after all. Instance segmentation approaches, such as~\cite{neven2018towards,pan2018}, are not subject to a specific segmentation order, resulting in a slightly higher accuracy in table~\ref{tab:sota}. An exciting next step is to combine our differentiable line fitting module with an approach to output each line-instance sequentially. Hence, we can combine the best of both methods. However this is outside the scope of this paper.

\section{Conclusion}
\label{sec:conclusion}
In this work we proposed a method for estimating lane curvature parameters by solving a weighted least-squares problem in-network, where the weights are generated by a deep network conditioned on the input image. The network was trained to minimize the area between the predicted lane lines and the ground truth lane lines, using a geometric loss function. We visualized the dynamics of backpropagating through a weighted least-squares fitting procedure, and provided an experiment on a 
lane detection task showing that our end-to-end method outperforms a two-step procedure despite the weaker supervision signal. Our end-to-end approach shows clear improvement operating at 70 fps.
The general idea of backpropagating through an in-network optimization step could prove effective in other computer vision tasks as well, for example in the framework of active contour models. In such a setting, the least-squares fitting module could perhaps be replaced with a more versatile differentiable gradient descent module. This will be explored in future work.

\textbf{Acknowledgement:} This work was supported by Toyota, and was partially carried out at the TRACE Lab at KU Leuven (Toyota Research on Automated Cars in Europe).


{\small
\bibliographystyle{ieee}
\bibliography{egbib}
}

\end{document}